\documentclass[sigconf]{acmart}

\AtBeginDocument{%
  }
\usepackage{multirow} 
\usepackage{pifont}
\usepackage{algorithm}
\usepackage{algpseudocode}
\usepackage{amsmath}
\usepackage{xcolor}

\renewcommand\footnotetextcopyrightpermission[1]{}
\definecolor{softgreen}{RGB}{0,154,85} %
\definecolor{softred}{RGB}{250,120,130}

\begin{document}

\title{SDVPT: Semantic-Driven Visual Prompt Tuning for Open-World Object Counting}

\author{Yiming Zhao}
\email{zhaoyiming23@mails.ucas.ac.cn}
\affiliation{%
  \institution{UCAS}
  \city{Beijing}
  \country{China}}

\author{Guorong Li}
\authornote{Guorong Li is the corresponding author.}
\email{liguorong@ucas.edu.cn}
\affiliation{%
  \institution{UCAS}
  \city{Beijing}
  \country{China}}

\author{Laiyun Qing}
\email{lyqing@ucas.ac.cn}
\affiliation{%
  \institution{UCAS}
  \city{Beijing}
  \country{China}}

\author{Amin Beheshti}
\email{amin.beheshti@mq.edu.au}
\affiliation{%
  \institution{Macquarie University}
  \city{Sydney}
  \country{Australia}}

\author{Jian Yang}
\email{jian.yang@mq.edu.au}
\affiliation{%
  \institution{Macquarie University}
  \city{Sydney}
  \country{Australia}}

\author{Michael Sheng}
\email{michael.sheng@mq.edu.au}
\affiliation{%
  \institution{Macquarie University}
  \city{Sydney}
  \country{Australia}}

\author{Yuankai Qi}
\email{yuankai.qi@mq.edu.au}
\affiliation{%
  \institution{Macquarie University}
  \city{Sydney}
  \country{Australia}}

\author{Qingming Huang}
\email{qmhuang@ucas.ac.cn}
\affiliation{%
  \institution{UCAS}
  \city{Beijing}
  \country{China}}

\begin{abstract}
Open-world object counting leverages the robust text-image alignment of pre-trained vision-language models (VLMs) to enable counting of arbitrary categories in images specified by textual queries. 
However, widely adopted naive fine-tuning strategies concentrate exclusively on text-image consistency for categories contained in training, which leads to limited generalizability for unseen categories.
In this work, we propose a plug-and-play Semantic-Driven Visual Prompt Tuning framework (SDVPT) that transfers knowledge from the training set to unseen categories with minimal overhead in parameters and inference time.
First, we introduce a two-stage visual prompt learning strategy composed of Category-Specific Prompt Initialization (CSPI) and Topology-Guided Prompt Refinement (TGPR). The CSPI generates category-specific visual prompts, and then TGPR distills latent structural patterns from the VLM’s text encoder to refine these prompts.
During inference, we dynamically synthesize the visual prompts for unseen categories based on the semantic correlation between unseen and training categories, facilitating robust text-image alignment for unseen categories.
Extensive experiments integrating SDVPT with all available open-world object counting models demonstrate its effectiveness and adaptability across three widely used datasets: FSC-147, CARPK, and PUCPR+. 
Code is available \href{https://github.com/Eamon-0v0/SDVPT}{here}
\end{abstract}


\received{20 February 2007}
\received[revised]{12 March 2009}
\received[accepted]{5 June 2009}

\maketitle

\section{Introduction}

\setlength{\textfloatsep}{10pt}  
\begin{figure}[!t]  
\centering  
\subfloat[Open-world Object Counting]{\includegraphics[width=0.45\textwidth]{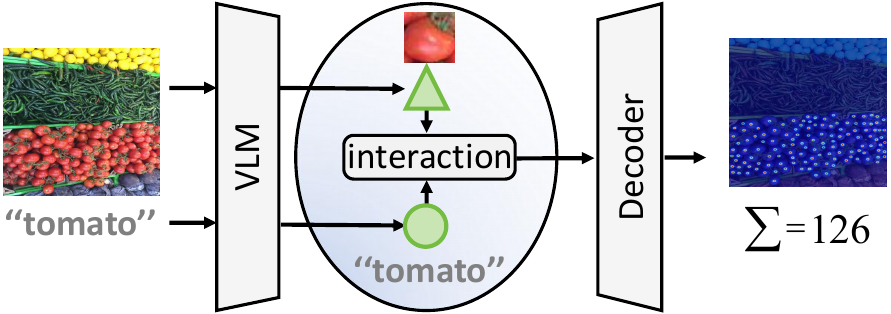}%
\label{FSC}}  
  
\subfloat[Results of Naive Fine-tuning Strategies]{\includegraphics[width=0.45\textwidth]{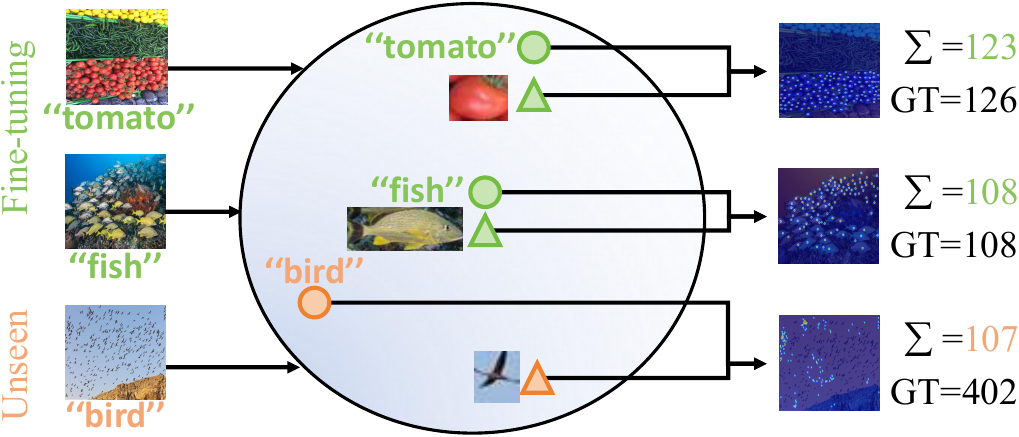}%
\label{OWC}}  
\caption{Illustration of the open-world object counting pipeline and limitations of naive fine-tuning strategies. (a) Open-world object counting depends on text-image alignment to enable user interaction and decoding. (b) Naive fine tuning and visual prompt tuning strategies neglect text-image alignment for unseen categories, resulting in inaccurate predictions during testing.}
\label{fig:1}  
\end{figure}

Object counting aims to enumerate objects within dense scenes~\cite{chan2008privacy}. Traditional class-specific approaches count predefined categories (e.g., crowds~\cite{boominathan2016crowdnet,shang2016end}, vehicles~\cite{mundhenk2016large}, animals~\cite{arteta2016counting}) using specialized models. 
Few-shot object counting~\cite{ranjan2021learning} extends the counting capability to arbitrary classes specified by user-provided annotated exemplars. However, its reliance on precise annotations of exemplars and complex interaction workflows restricts real-world applicability~\cite{jiang2023clip,kang2024vlcounter}. 
Recently, open-world object counting~\cite{Xu_2023_CVPR} has received increasing attention, which enables the counting of target objects beyond training set categories by using only textual prompts, thus providing a more flexible and scalable solution.

Existing open-world object counting approaches commonly adopt the pipeline illustrated in Fig.~\ref{fig:1}(a), leveraging the robust image-text alignment of pre-trained vision-language models (VLMs) to map image and text into a joint embedding space, where interaction is first performed to compute the image-text similarity map, followed by decoding.
Given that object counting encounters dense scenes differing from those in VLM pre-training, fine-tuning VLMs is essential~\cite{AminiNaieni23}. 
Existing studies~\cite{jiang2023clip,kang2024vlcounter} have shown that, due to text similarity with pre-training and dense scenarios differing from pre-training, fine-tuning the text encoder yields negligible benefits, whereas fine-tuning the visual encoder with full tuning or visual prompts shows promising effectiveness.
However, naive fine-tuning strategies, such as full tuning or visual prompt tuning, concentrate exclusively on text-image consistency for categories contained in training data~\cite{zhou2022cocoop}. This leads to limited generalizability for unseen categories, as shown in Fig.~\ref{fig:1}(b).

To address the above-mentioned issues, we propose a plug-and-play Semantic-Driven Visual Prompt Tuning framework (SDVPT), which transfers knowledge from training categories to unseen ones by aligning visual prompt topologies with text embeddings. 
Specifically, we design a two-stage visual prompt learning strategy consisting of Category-Specific Prompt Initialization (CSPI) and Topology-Guided Prompt Refinement (TGPR).
The CSPI adapts the visual-text correspondence of pre-trained VLMs to counting tasks by learning category-specific visual prompts through contrastive instance-text alignment, ensuring precise localization and density-aware feature extraction for seen classes.
To improve the generalization ability of the learned visual prompt, we design an aggregation strategy for visual prompts in TGPR that explicitly models topological relationships in the pre-trained VLM’s text embedding space. By distilling latent structural patterns from the text encoder, TGPR transfers these topology-aware constraints to refine visual prompts, effectively bridging the modality gap while preserving semantic consistency.
During inference, visual prompts for unseen-class are dynamically synthesized via aggregation of semantic related seen-class prompts, enabling zero-shot counting without architectural changes. This two-stage design preserves VLM knowledge while ensuring cross-category topological consistency.

As a plug-and-play framework, we integrate the proposed method with all available VLM-based open-world object counting models (i.e., CLIP-Count~\cite{jiang2023clip}, VLCounter~\cite{kang2024vlcounter}, CounTX~\cite{AminiNaieni23}, CountGD~\cite{amini2024countgd}), 
and test them on the FSC-147~\cite{ranjan2021learning}, CARPK~\cite{hsieh2017drone}, and PUCPR+~\cite{hsieh2017drone} datasets. Experimental results demonstrate effectiveness of our method, with the CountGD-integrated version achieving new state-of-the-art open-world object counting results across all datasets.

In summary, this work presents the following main contributions:
\begin{itemize}
\item We propose a plug-and-play Semantic-Driven Visual Prompt Tuning
framework (SDVPT) for open-world object counting 
with minimal overhead in parameters and inference time, which is composed of a two-stage visual prompt learning, i.e., Category-Specific Prompt Initialization and Topology-
Guided Prompt Refinement. 
\item The Topology-
Guided Prompt Refinement is designed to distill latent structural patterns from VLM's text encoder to refine visual prompts, enabling them to synthesize visual prompts for unseen categories based on semantic correlation between categories.
\item Extensive experiments combining the proposed SDVPT with four open-world object counting methods across three datasets demonstrate its favorable effectiveness.
\end{itemize}

\section{Related Works}

\subsection{Object Counting}
\noindent\textbf{Class-Specific Object Counting} aims to count objects belonging to a specific class. Existing approaches are broadly categorized into detection-based methods~\cite{song2021rethinking, liang2022end, sam2020locate, deng2023improving} and regression-based methods~\cite{shang2016end, sindagi2017generating, lempitsky2010learning, song2021choose, ma2019bayesian}. 
Detection-based methods tally objects using bounding box predictions but often falter in scenarios with high object density. 
To address this limitation, Lempitsky et al.\cite{lempitsky2010learning} pioneered the use of density maps to model spatial object distributions, a paradigm that has gained widespread adoption. 
Subsequent advancements have enhanced density map quality through contextual encoding\cite{sindagi2017generating}, multi-scale architectures~\cite{song2021choose}, and novel loss functions~\cite{ma2019bayesian}. 
Nevertheless, these methods typically demand extensive annotated training data and are constrained to specific classes, limiting their generalizability due to inherent specialization and complexity.

\noindent\textbf{Few-Shot Object Counting} aims to enumerate objects of any target class in a query image by specifying the category with a few exemplars.
GMNNet~\cite{lu2019class} pioneered object recognition through exemplar-to-image similarity matching. 
Subsequent methods, such as FamNet~\cite{ranjan2021learning}, CFOCNet~\cite{yang2021class}, BMNet~\cite{shi2022represent}, and SAFECount~\cite{you2023few}, refined similarity modeling within this paradigm.
Later approaches, including SPDCN~\cite{lin2022scale}, LOCA~\cite{djukic2022low}, and DAVE~\cite{pelhan2024dave}, further explored exemplar interactions to enhance guidance.
More recently, advanced pretrained visual backbones, such as ViT~\cite{dosovitskiy2020vit} and GroundingDINO~\cite{liu2024grounding}, have been utilized for robust feature extraction, exemplified by CACVIT~\cite{wang2024vision} and CountGD~\cite{amini2024countgd}.
Despite generalizing object counting to arbitrary categories, this few-shot paradigm still relies on costly bounding box-level annotations. Moreover, in real-world settings, requiring new exemplar annotations for each image undermines the interactive experience for human users.

\begin{figure*}[!t]
\centerline{\includegraphics[width=0.9\textwidth]{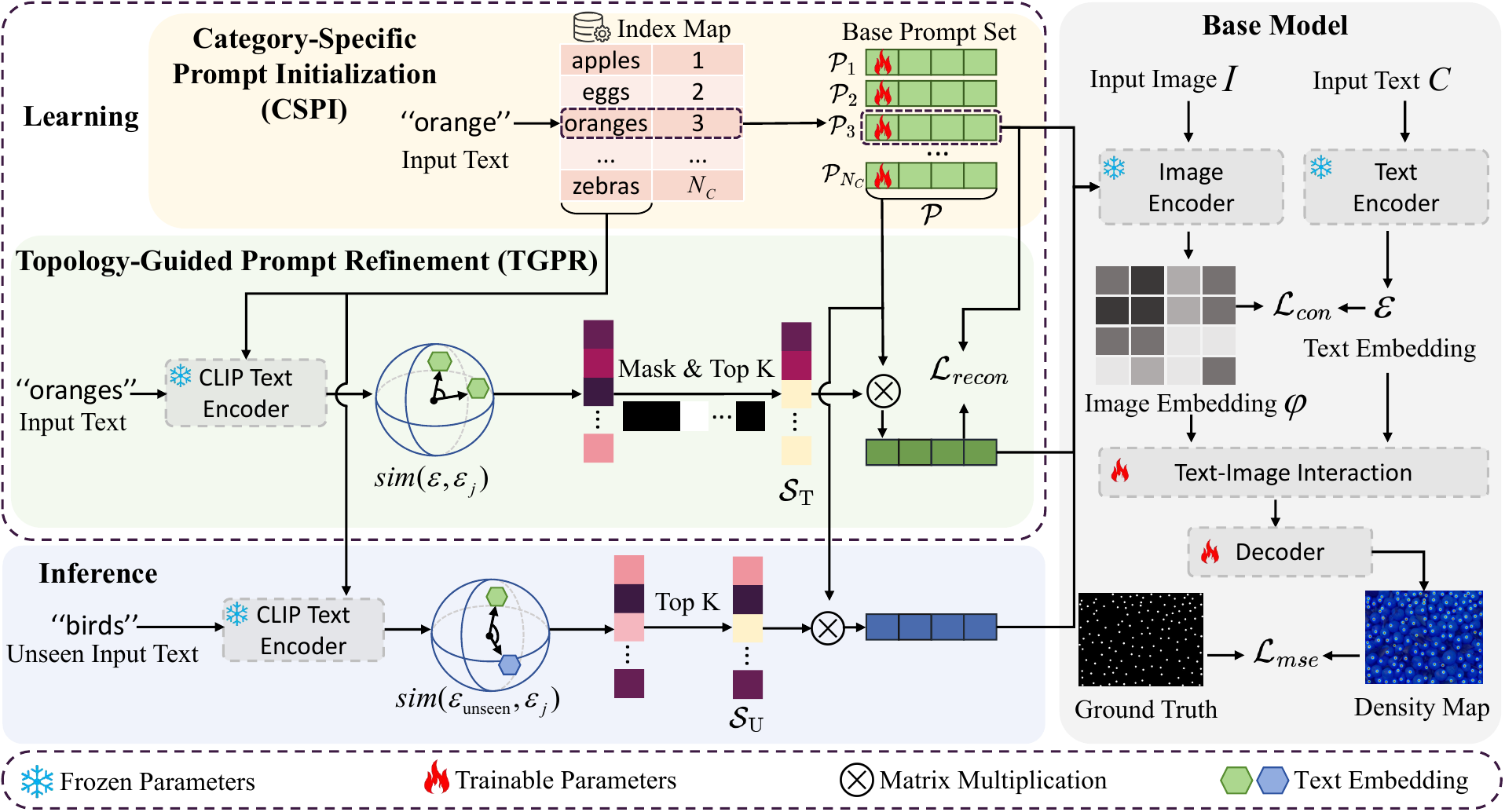}}
\caption{{\bf Main architecture of the proposed method.} Our method comprises CSPI and TGPR modules. The CSPI in
Sec.~\ref{CSPI} trains a set of category-specific visual prompts, while the TGPR in Sec.~\ref{TGPR} transfers the topological structure of text embeddings onto them. 
For inference, we employ an aggregation strategy similar to TGPR to harness the topological structure of text embeddings spanning unseen and training categories, thereby extending knowledge from the training set to unseen categories.
}
\label{fig:2}
\end{figure*}

\noindent\textbf{Open-World Object Counting}, also known as zero-shot object counting, was initially introduced by Xu \textit{et al.}\cite{Xu_2023_CVPR}.
Their method first detects visual exemplars based on textual information.
Then, it uses the exemplars as inputs for existing few-shot object counting models, such as FamNet\cite{ranjan2021learning}.
CounTX~\cite{AminiNaieni23} leverages the extensive pre-training knowledge of VLMs to map image and text inputs into a joint embedding space, enabling end-to-end counting for the first time.
Unlike CounTX, which fully optimizes CLIP~\cite{radford2021learning}’s image encoder, CLIP-Count~\cite{jiang2023clip} and VLcounter~\cite{kang2024vlcounter} employ Visual Prompt Tuning (VPT)~\cite{jia2022visual} to achieve lighter and more efficient adaptation.
CLIP-Count introduces a training framework that synergistically fine-tunes CLIP using visual and textual prompts.
VLCounter investigates the interaction between visual prompts and text embeddings, proposing a semantic-conditioned prompt tuning mechanism.
CountGD~\cite{amini2024countgd} employs the state-of-the-art vision-language model GroundingDINO~\cite{liu2024grounding} as its backbone to achieve superior performance without fine-tuning.
Although these methods acknowledge the importance of fine-tuning for VLMs, their adopted strategies neglect image-text consistency for unseen categories.

\vspace{-0.5em}
\subsection{Prompt Tuning}
Prompt Tuning, initially proposed in natural language processing (NLP), enables large pre-trained models to efficiently adapt to downstream tasks with a minimal memory footprint.
CoOp~\cite{zhou2022coop} extends prompt tuning to multimodal tasks by substituting learnable vectors for the context words in CLIP’s text encoder.
Visual Prompt Tuning (VPT)~\cite{jia2022visual} further adapts this approach to the visual modality, embedding learnable vectors into the context of a transformer-based image encoder.

In open-world object counting, the complex scenes differing from those encountered during VLMs pre-training have driven existing research to seek effective fine-tuning strategies. 
CounTX~\cite{AminiNaieni23} reveals that fine-tuning the text encoder yields limited improvements, whereas fine-tuning the visual encoder proves highly effective. 
Meanwhile, CLIP-Count~\cite{jiang2023clip} further demonstrates that prompt tuning outperforms full tuning for this task. 
Subsequent studies, such as VLCounter~\cite{kang2024vlcounter}, have adopted the approach of fine-tuning with visual prompts.

However, Zhou \textit{et al.}~\cite{zhou2022cocoop} and Gan \textit{et al.}~\cite{gan2023decorate} have identified that prompt tuning is prone to overfitting, as it focuses exclusively on text-image alignment for training categories, resulting in diminished generalization to unseen categories.
Unfortunately, there is still no effective way to address this challenge in the open-world counting task.
In contrast, several related studies have been proposed in other tasks. For instance, in image classification, CoCoOp~\cite{zhou2022cocoop} and CoPL~\cite{goswami2024copl} employ a meta net to encode visual features, incorporating them into textual prompts to enable generalization to unseen categories. 
However, our experimental results corroborate the findings of CounTX~\cite{amini2024countgd}, demonstrating that fine-tuning the text encoder with these approaches in open-world object counting produces only limited improvements.

Building on these efforts, we propose a novel VPT method for open-world object counting that leverages semantic information from pre-trained VLMs to generalize knowledge learned from training categories to unseen categories, mitigating the overfitting limitations of traditional VPT.

\section{Method}
In this section, we first provide the definition for open-world object counting and the VLMs. 
Subsequently, we introduce the two-stage visual prompt learning framework comprising Category-Specific Prompt Initialization (CSPI) in Sec.~\ref{CSPI} and Topology-Guided Prompt Refinement (TGPR) in Sec.~\ref{TGPR}. 
Finally, we depict the pipelines for inference.
The overview of our framework is shown in Fig.~\ref{fig:2}.

\vspace{1mm}
\noindent\textbf{Problem Formulation}
Open-world object counting aims to enumerate objects of a specified class $C$ within an image $I \in \mathbb{R}^{H \times W \times 3}$. The training set containing $\mathbb{N}$ samples with $N_C$ categories is defined as $\mathcal{D}_{\mathrm{train}}=\{(I_{i},C_{i}^{train},D_{i})\}_{i=1}^{i=\mathbb{N}}$, where $D_i$ represents the density map of objects belonging to the image $I_{i}$, $C_i^{train}$ denotes the category name corresponding to one of the $N_C$ categories. The test set $\mathcal{D}_{\mathrm{test}}=\{(I_{i},C_{i}^{test},D_{i})\}_{i=1}^{i=\mathbb{M}}$ contains entirely different semantic categories from the training set, i.e. $C^{train}\cap C^{test}=\varnothing$.

\vspace{1mm}
\noindent\textbf{Preliminary: Vision-language models}
Vision-language models (VLMs), such as CLIP~\cite{radford2021learning} and GroundingDINO~\cite{liu2024grounding}, comprise a text encoder $\phi_{T}(\cdot)$ and an transformer-based image encoder $\phi_{I}(\cdot)$. Through pre-training with contrastive learning, these models map the image $I_{i}$ and the text $C_{i}^{train}$ belonging to the $k^{th}$ category into a joint embedding space, producing image embedding $\varphi_i$ and text embeddings $\varepsilon_k$ that are close to each other. The contrastive loss is formally defined as:
\begin{equation}
\small
\mathcal{L}_{\text{con}}=-(\log\frac{\exp(\frac{s(\varphi_{i},\varepsilon_{k})}{\tau})}{\sum\nolimits_{j=1}^{N_C}(\exp(\frac{s(\varphi_{i},\varepsilon_{j})}{\tau})}+\log\frac{\exp(\frac{s(\varphi_{i},\varepsilon_{k})}{\tau})}{\sum\nolimits_{j=1}^{N}(\exp(\frac{s(\varphi_{j},\varepsilon_{k})}{\tau})}),
\label{eq:con_loss}
\end{equation}
where $s(\cdot)$ denotes a similarity metric (e.g., cosine similarity), $\tau$ is a temperature parameter, $N$ is the batch size, and all other images and texts in the batch do not belong to the $k$-th category.

Due to the difference in visual scenarios between open-world object counting and the pre-training of VLMs, directly applying pre-trained VLMs cannot reliably ensure the alignment between text and image embeddings. Therefore, both our method and mainstream approaches continue to employ a similar contrastive loss during the fine-tuning.

\begin{figure}[t]
\centerline{\includegraphics[width=0.48\textwidth]{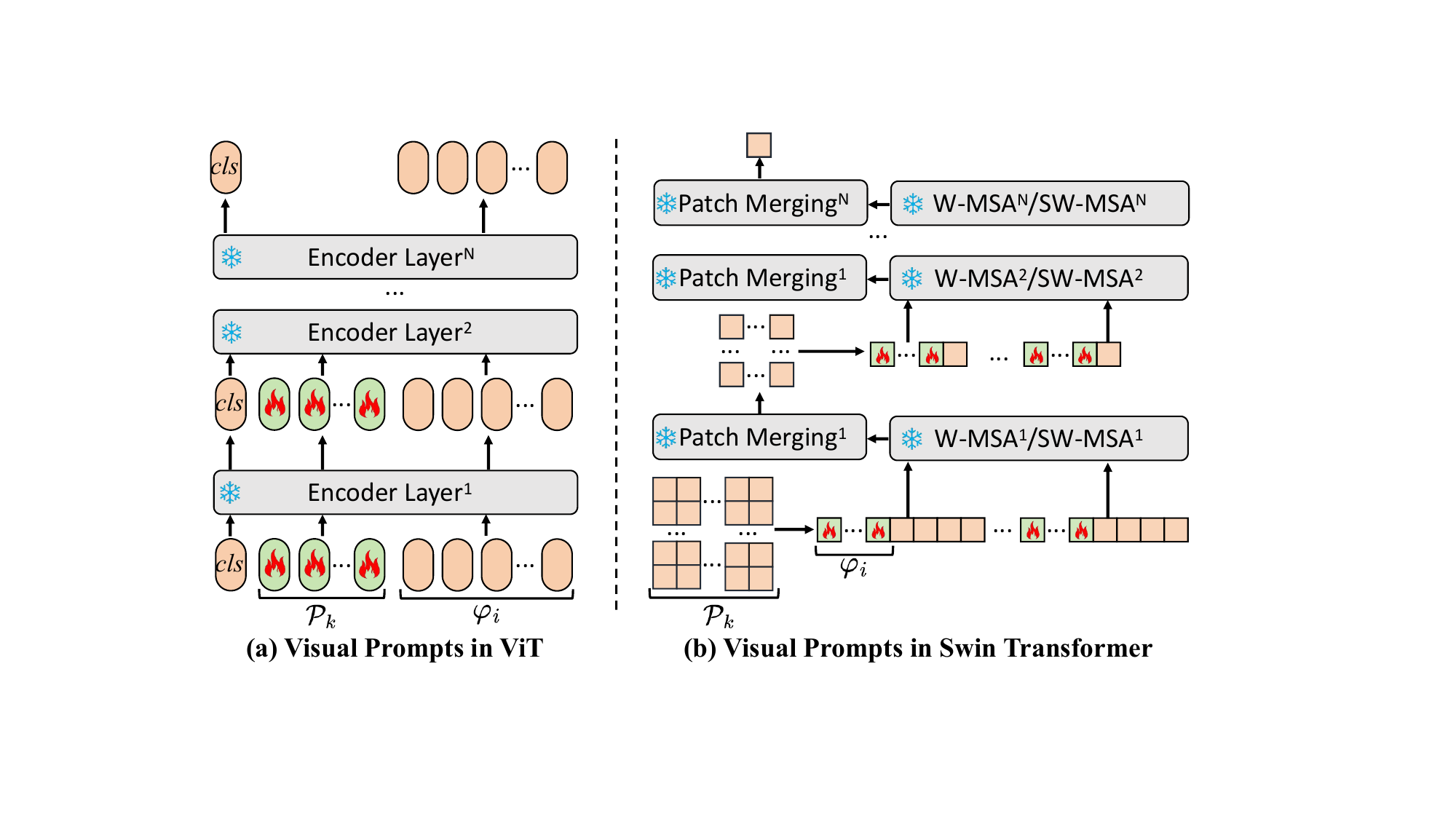}}
\caption{Illustration of visual prompt integration. (a) For ViT, we embed the prompt between the $cls$ token and the image embedding. (b) For Swin Transformer, we insert the visual prompt before Window Multi-Head Self-Attention (W-MSA) and Shifted Window Multi-Head Self-Attention (SW-MSA), removing it during the patch merging stage.}
\label{fig:3}
\end{figure}

\subsection{Category-Specific Prompt Initialization }
\label{CSPI}
CSPI aims to adapt VLMs to counting tasks by learning class-specific visual prompts. 
For the image $I_i$ and the text $C_{i}^{train}$ belonging to the $k^{th}$ category, it fine-tunes the image encoder $\phi_I(\cdot)$ using a category-specific visual prompt $\mathcal{P}_k$, thereby infusing the   prompts with category-specific knowledge:
\begin{equation}
\varphi_i=\phi_I(I_i,\mathcal{P}_k).
\end{equation}

Specifically, $\mathcal{P}_k$ consists of the prompt for several layers of the image encoder, i.e., $\mathcal{P}_k=[\mathcal{P}_k^{1},\mathcal{P}_k^{2},...,\mathcal{P}_k^{L}]$, where $L$ represents the number of visual prompt layers. 
All category-specific visual prompts collectively form the base prompt set $\mathcal{P}=[\mathcal{P}_{1},\mathcal{P}_{2},...,\mathcal{P}_{N_C}]$, where $N_C$ is the number of training set categories.
During training, we fine-tune the $\mathcal{P}$ while keeping both the text encoder and the image encoder frozen.

For a ViT-based $\phi_{I}(\cdot)$, as shown in Fig.~\ref{fig:3}(a), we integrate the visual prompt 
between the $cls$ token and image embedding. 
The process for the $l^{th}$ layer is formulated as:
\begin{equation}
[ cls, \_, \varphi^{l+1}_i ]=\phi_{I}^{l}([cls, \mathcal{P}^{l}_k, \varphi^{l}_i]),
\end{equation}
where $\_$ denotes removing the token at the position corresponding to $\mathcal{P}^{l}_k$ in the output of the $l^{th}$ layer, followed by inserting $\mathcal{P}^{l+1}_k$ before feeding $\varphi^{l+1}_i$ into the $(l+1)^{th}$ layer.

For the Swin Transformer-based $\phi_{I}(\cdot)$, we incorporate the visual prompts within local windows, excluding them during patch merging, as shown in Fig.~\ref{fig:3}(b).

As a plug-and-play framework, when integrated with   existing open-world counting model, the loss function of CSPI is 
comprised of three essential components:
the model's original loss $\mathcal{L}_{\mathrm{model}}$ to maintain its inherent capabilities, a contrastive loss $\mathcal{L}_{\mathrm{con}}$ for cross-modal alignment, and a mean squared error loss $\mathcal{L}_{\mathrm{mse}}$ for count supervision. The composite loss is formally defined as:
\begin{equation}
\mathcal{L}_{\mathrm{CSPI}}=\mathcal{L}_{\mathrm{mse}}+\lambda_1\mathcal{L}_{\mathrm{con}}+\lambda_2\mathcal{L}_{\mathrm{model}},
\label{Loss_CSPI}
\end{equation}
where $\lambda_1$ and $\lambda_2$ denote weighting coefficients optimized through grid search and $\mathcal{L}_{mse}$ imposes direct count supervision through count-level regression:
\begin
{equation}\mathcal{L}_{\mathrm{mse}}=\frac{1}{N}\sum_{i=1}^N({Y}_{pred}^{i}-Y_{gt}^{i})^2,
\end{equation}
where ${Y}_{gt}^{i}$ represents the ground-truth count, ${Y}_{pred}^{i}$ denotes the predicted count, $N$ is batch size. 

The contrastive loss $\mathcal{L}_{\text{con}}$ in Eq.~\eqref{eq:con_loss}, designed to align text and visual embeddings, typically adopts the baseline model’s default implementation.
For example, in CLIP-Count~\cite{jiang2023clip}, $\mathcal{L}_{\text{con}}$ leverages patch embeddings instead of the $cls$ token-derived image embedding, a convention we adopt accordingly.

\begin{figure}[t]
\centerline{\includegraphics[width=0.45\textwidth]{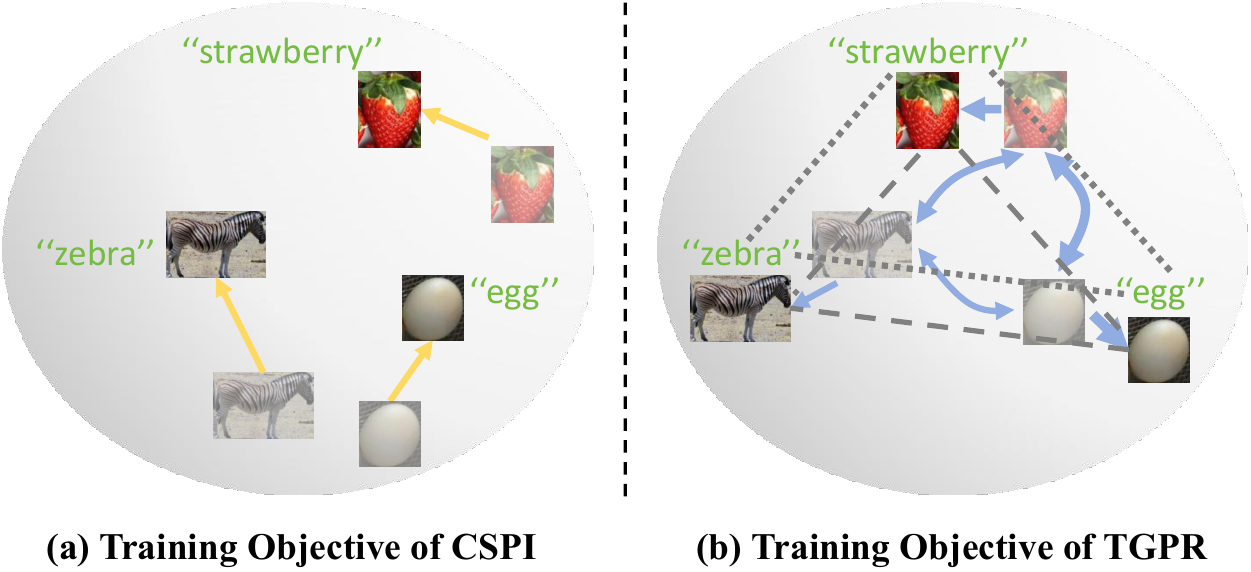}}
\caption{Illustration of the learning objectives for CSPI and TGPR. (a) CSPI enforces alignment between text and visual embeddings within categories, disregarding the topological structure among visual embeddings. (b) TGPR harmonizes the topological structure of visual embeddings with that of text embeddings.}
\label{fig:4}
\end{figure}

\subsection{Topology-Guided Prompt Refinement}
\label{TGPR}

While CSPI effectively learns visual prompts to known categories, its class-specific design struggles to generalize to unseen counting targets.
To overcome this limitation, we propose Topology-Guided Prompt Refinement (TGPR) for the second-stage visual prompt learning. 
As shown in Fig.~\ref{fig:4}(b), TGPR transfers the topological structure among the text embeddings of the training set to the base visual prompt $\mathcal{P}$ set while preserving their counting abilities learned in CSPI.

Unlike CSPI, which employs category-specific visual prompts, we introduce an aggregation strategy for the base prompt set to generate a visual prompt tailored to the current category.
Specifically, for the image $I_i$ and its corresponding text input $C_{i}^{train}$ of the $k^{th}$ category, e.g., the "oranges" in Fig.~\ref{fig:2}, we posit that the topological relationship between the text embedding $\varepsilon_k$ and those of other training categories can serve as a proxy for the relevance between the optimal visual prompt for the $k^{th}$ category and other visual prompts.
Leveraging this, we utilize the similarities between the text embedding $\varepsilon_k$ and those of other training categories as prior constraints to fuse the base prompt set $\mathcal{P}$, thereby producing a visual prompt customized for the $k^{th}$ category:
\begin{equation}
\label{eq_st}
\varphi_i=\phi_I(I_i,\sum_{j\in\mathcal{S}_{\text{T}}}sim(\varepsilon_k,\varepsilon_{j})\cdot\mathcal{P}_j),
\end{equation}
where $sim(\cdot)$ denotes cosine similarity, $j\in\mathcal{S}_{\text{T}}$ indicates the selection of the top $K$ text embeddings most similar to $\varepsilon_{k}$.
This top $K$ filtering strategy effectively mitigates the detrimental impact of highly dissimilar categories. The $\mathcal{S}_{\text{T}}$ is formulated as:
\begin{equation}\mathcal{S}_{\mathrm{T}}=\arg\mathrm{topK}_{j\in\{1,\ldots,N_C\},j\neq k}\mathrm{sim}(\varepsilon_k,\varepsilon_j),\end{equation}
where $N_C$ is the number of training categories, $j\neq k$ represents a self-exclusion mask during the top-$K$ selection to exclude the similarity of the current category with itself, ensuring the fused prompt leverages complementary information from other categories.

Additionally, we introduce a reconstruction loss to optimize the parameters of all prompts involved in the fusion, measured as the $L_2$ distance between the fused prompt and the category-specific prompt $\mathcal{P}_k$. This loss encourages a balance between seeking an optimal topological structure and achieving optimal counting performance, formulated as:
\begin{equation}\mathcal{L}_{\mathrm{recon}}=\|\mathcal{P}_k-\sum_{j\in\mathcal{S}_{\text{T}}}sim(\varepsilon_k,\varepsilon_{j})\cdot\mathcal{P}_j\|_2^2 .
\end{equation}

Consequently, the total loss in TGPR is :
\begin{equation}
\label{loss_tgpr}
\mathcal{L}_{\mathrm{TGPR}}=\mathcal{L}_{\mathrm{mse}}+\lambda_1\mathcal{L}_{\mathrm{con}}+\lambda_2\mathcal{L}_{\mathrm{model}}+\lambda_3\mathcal{L}_{\mathrm{recon}} .
\end{equation}

The detailed learning process is shown in Algorithm~\ref{alg:SDVPT}.

\subsection{Inference}
\label{Inference}
When addressing unseen categories,  similar to aggregation strategy in TGPR, we select visual prompts of semantic related category to dynamically synthesize the visual prompt for the current unseen category:
\begin{equation}
\mathcal{P}_{\text{unseen}}=\mathop{\sum}_{j\in\mathcal{S}_\text{U}}sim(\varepsilon_{\text{unseen}},\varepsilon_{j})\cdot\mathcal{P}_j.
\end{equation}
where $\mathcal{S}_{\text{U}}=\arg\mathrm{topK}_{j\in\{1,\ldots,N_C\}}sim(\varepsilon_{\text{unseen}},\varepsilon_j)$.
In this way, we align the image embeddings of unseen categories with their corresponding text embeddings in the joint embedding space, thus ensuring their consistency.

\begin{algorithm}[h]
\caption{Learning process of SDVPT}
\label{alg:SDVPT}
\begin{algorithmic}[1]
\Require Training set $\mathcal{D}_{\mathrm{train}}$, test set $\mathcal{D}_{\mathrm{test}}$, base prompt set $\mathcal{P}$, image encoder $\phi_I(\cdot)$, text encoder $\phi_T(\cdot)$, counting model $M$
\Ensure  Final model $M_{final}$

\Comment{\textbf{Learning Stage 1: CSPI}}
\State Load and freeze pretrained $\phi_I(\cdot)$, $\phi_T(\cdot)$, initialize $M$, $\mathcal{P}$
\State Define index map $Map$ from $\mathcal{D}_{\mathrm{train}}$
\For{$e = 1$ to $E_1$}
    \For{batch $(I_{i},C_{i}^{train},D_{i})$ in $\mathcal{D}_{\mathrm{train}}$}
        \State $\mathcal{P}_k=Map(C_{i}^{train},\mathcal{P})$
        \Comment{Select prompt $\mathcal{P}_k$ from $\mathcal{P}$}
        \State update $M$, $\mathcal{P}_k$ using $\mathcal{L}_{\mathrm{CSPI}}$ defined by Eq.~(\ref{Loss_CSPI})
    \EndFor
\EndFor

\Comment{\textbf{Learning Stage 2: TGPR}}
\State $\varepsilon_{\mathrm{train}} = \phi_T(C^{\mathrm{train}})$ \Comment{$C^{\mathrm{train}}$: all classes from $\mathcal{D}_{\mathrm{train}}$}
\For{$e = E_1 + 1$ to $E_2$}
    \For{batch $(I_{i},C_{i}^{train},D_{i})$ in $\mathcal{D}_{\mathrm{train}}$}
        \State $\varepsilon_k = \phi_T(C_{i}^{train})$
        \State $\mathcal{S}_{\mathrm{T}}=\arg\mathrm{topK}_{j\in\{1,\ldots,N_{C}\},j\neq k}\mathrm{sim}(\varepsilon_k,\varepsilon_j)$
        \State
        $\mathcal{P}_\mathrm{TGPR}=\sum_{j \in \mathcal{S}_{\mathrm{T}}} \mathrm{sim}(\varepsilon_k, \varepsilon_j) \cdot \mathcal{P}_j$
        \State update $M$, $\mathcal{P}$ using $\mathcal{L}_{\mathrm{TGPR}}$ defined by Eq.~(\ref{loss_tgpr})
    \EndFor
\EndFor

\State $M_{final} \gets M$ \Comment{Save the final model}
\State \Return $M_{final}$, $R$
\end{algorithmic}
\end{algorithm}

\section{Experiments}

\subsection{Experimental Details}

\noindent\textbf{Implementation Details.} 
We integrated our framework with four base models: CLIP-Count~\cite{jiang2023clip}, VLCounter~\cite{kang2024vlcounter}, CounTX~\cite{AminiNaieni23}, and CountGD~\cite{amini2024countgd}. For all experiments, we follow the same input size, data augmentation, learning rate, optimizer, and category name format as the integrated methods, with specific details available in their public papers or codebases.
For CLIP-Count and VLCounter, we also adopted the same visual prompt size and layer selection as the original methods, while for CounTX and CountGD, the number of visual prompt tokens was set to 10 and 5, respectively, applied across all image encoder layers.
The additional hyperparameters introduced by SDVPT include the number of epochs for CSPI and TGPR, the top-$K$ selection in Eq.~(\ref{eq_st}), and the loss function weights in Eq.~(\ref{loss_tgpr}), with integration details for each model provided in Tab.~\ref{tab:2}.

\setlength{\textfloatsep}{1pt}  
\begin{table}[h]
\vspace{-2mm}
\small
\begin{center}
\caption{Additional hyperparameters when integrated with different base models. $\lambda_2 = 0$ indicates that the base model has no additional loss $\mathcal{L}_{\mathrm{model}}$.} 
\label{tab:2}
\setlength{\tabcolsep}{4pt}
\begin{tabular}{cccccc}
  \toprule
  
\multirow{2}{*}{\shortstack{Base \\ Model}} & 
  \multirow{2}{*}{\shortstack{CSPI \\ Epoch}} & 
    \multirow{2}{*}{\shortstack{TGPR \\ Epoch}} & 
    \multirow{2}{*}{top-$K$} &
    \multirow{2}{*}{$(\lambda_1,\lambda_2,\lambda_3)$}&
    \multirow{2}{*}{GPU} \\
    \\

  \midrule
CLIP-Count~\cite{jiang2023clip} & 100 & 200 & 70 & (1,0,10) &RTX 3090 \\

VLCounter~\cite{kang2024vlcounter} & 50 & 350 & 70 & (1e-6,0,1) &RTX 3090 \\

CounTX~\cite{AminiNaieni23} & 100 & 400 & 80 & (1,0,0.1) & RTX A6000 \\

CountGD~\cite{amini2024countgd}& 5 & 30 & 70 & (1,1,1) & RTX A6000 \\

   \bottomrule
\end{tabular}
\end{center}
\end{table}

\begin{table*}[!t]
\small
\begin{center}
\caption{Quantitative performance on the FSC-147 dataset. (*) denotes our reproduced results. Full Tuning means freezing the text encoder and fully fine-tuning the image encoder. The best results are highlighted in bold.} 
\label{tab:tab1}

\setlength{\tabcolsep}{1pt}
\begin{tabular}{ccccccccccc}
  \toprule

  \multirow{2}{*}{Methods} & 
  \multirow{2}{*}{Source} & 
    \multirow{2}{*}{Tuning} & 
  \multicolumn{4}{c}{Val Set}& \multicolumn{4}{c}{Test Set}\\
 \cmidrule{4-11}
  & & & MAE$\downarrow$ & RMSE$\downarrow$  & NAE$\downarrow$  &SRE$\downarrow$ & MAE$\downarrow$ & RMSE$\downarrow$  & NAE$\downarrow$ &SRE$\downarrow$ \\
  \midrule

CLIP-Count~\cite{jiang2023clip}  &MM 2023 & VPT &18.79& 61.18 & - & - & 17.78& 106.62 & - &- 
\\

CLIP-Count* & - & VPT & 19.14 & 64.81 & 0.38 & 4.06 & 17.33 & 106.82 & 0.37 & 9.41 \\ 
CLIP-Count*+Ours & - & SDVPT & 16.74{\color{softgreen} $\downarrow$12.54\%} & 55.80{\color{softgreen} $\downarrow$13.90\%} & 0.30{\color{softgreen} $\downarrow$21.05\%} & 3.43{\color{softgreen} $\downarrow$15.52\%} & 16.92{\color{softgreen} $\downarrow$2.37\%} & 101.77{\color{softgreen} $\downarrow$4.73\%} & 0.34{\color{softgreen} $\downarrow$8.11\%} & 9.20{\color{softgreen} $\downarrow$2.23\%} \\ 

  \midrule
  
VLCounter~\cite{kang2024vlcounter}  &AAAI 2024& VPT &18.06& 65.13 & - & - & 17.05& 106.16& - & -
\\

VLCounter* & - & VPT & 18.16 & 65.75 & 0.31 & 3.42 & 17.57 & 107.49 & 0.33 & 6.85 \\

VLCounter*+Ours & - & SDVPT & 17.74 {\color{softgreen} $\downarrow$2.31\%} & 63.98 {\color{softgreen} $\downarrow$2.69\%} & 0.28 {\color{softgreen} $\downarrow$9.68\%} & 3.21 {\color{softgreen} $\downarrow$6.14\%} & 16.33 {\color{softgreen} $\downarrow$7.06\%} & 104.25 {\color{softgreen} $\downarrow$3.01\%} & 0.28 {\color{softgreen} $\downarrow$15.15\%} & 6.63 {\color{softgreen} $\downarrow$3.21\%} \\

  \midrule
  
CounTX~\cite{AminiNaieni23}  &BMVC 2023& Full Tuning &17.10& 65.61 &-&-& 15.88& 106.29&-&-
\\

CounTX* & - & Full Tuning & 17.92 & 65.82 & 0.31 & 4.33 & 15.98 & 107.15 & 0.32 & 8.95 \\ 

CounTX*+Ours & - & SDVPT & 16.89 {\color{softgreen} $\downarrow$5.75\%} & 64.77 {\color{softgreen} $\downarrow$1.60\%} & 0.28 {\color{softgreen} $\downarrow$9.68\%} & 3.40 {\color{softgreen} $\downarrow$21.48\%} & 15.23 {\color{softgreen} $\downarrow$4.69\%} & 105.85 {\color{softgreen} $\downarrow$1.21\%} & 0.30 {\color{softgreen} $\downarrow$6.25\%} & 8.63 {\color{softgreen} $\downarrow$3.58\%} \\ 
  
  \midrule
  
CountGD~\cite{amini2024countgd}  &NeurIPS 2024&  Freeze &12.14& 47.51 & -&-& 12.98& 98.35& -&-
\\

CountGD* & - & Freeze & 11.27 & 45.62 & 0.15 & 2.21 & 13.96 & 114.72 & 0.18 & 3.21 \\ 

CountGD*+Ours & - & SDVPT & \textbf{9.52} {\color{softgreen} $\downarrow$15.53\%} & \textbf{37.02} {\color{softgreen} $\downarrow$18.85\%} & \textbf{0.12} {\color{softgreen} $\downarrow$20.00\%} & \textbf{1.92} {\color{softgreen} $\downarrow$13.12\%} & \textbf{11.41} {\color{softgreen} $\downarrow$18.27\%} & \textbf{85.50} {\color{softgreen} $\downarrow$25.47\%} & \textbf{0.14} {\color{softgreen} $\downarrow$22.22\%} & \textbf{2.49} {\color{softgreen} $\downarrow$22.43\%} \\ 

   \bottomrule
\end{tabular}
\end{center}
\end{table*}

\vspace{-2mm}
\noindent\textbf{Datasets.} 
We utilized FSC-147~\cite{ranjan2021learning}, a large-scale few-shot object counting dataset comprising 6135 images across 147 categories. Consistent with prior methods~\cite{jiang2023clip,kang2024vlcounter,AminiNaieni23,amini2024countgd}, we used only category names and images as inputs to enable an open-world object counting setup.
CARPK~\cite{hsieh2017drone} comprises 1,488 images containing 89,777 vehicles, captured from approximately 40 meters above four parking lots. PUCPR+~\cite{hsieh2017drone} consists of 125 images with 17,000 vehicles, exhibiting varied weather conditions, including rainy, cloudy, and sunny.
We employed CARPK and PUCPR+ for cross-dataset validation to demonstrate the generalizability of our framework.

\vspace{1mm}
\noindent\textbf{Evaluation Metrics.} 
Following prior studies~\cite {jiang2023clip,kang2024vlcounter,AminiNaieni23,amini2024countgd}, we evaluate the performance using mean absolute error (MAE) and root mean squared error (RMSE):
\begin{small}
\begin{equation}\begin{gathered}
\text{MAE}=\frac{1}{N_{I}}\sum_{i=1}^{N_{I}}|{Y}_{pred}^{i}-Y_{gt}^{i}|, 
\text{RMSE}=\sqrt{\frac{1}{N_{I}}\sum_{i=1}^{N_{I}}({Y}_{pred}^{i}-Y_{gt}^{i})^{2}}, 
\end{gathered}\end{equation}\end{small}%
where $N_{I}$ is the number of images in the testing set, ${Y}_{pred}$ and ${Y}_{gt}$ are the predicted and ground truth object count, respectively.

Additionally, we adopt the Normalized Relative Error (NAE) and Squared Relative Error (SRE), which are normalized by the ground truth object count to mitigate the impact of extreme predictions:
\begin{small}
\begin{equation}\begin{gathered}
\text{NAE}=\frac{1}{N_{I}}\sum_{i=1}^{N_{I}}\frac{|{Y}_{pred}^{i}-{Y}_{gt}^{i}|}{{Y}_{gt}^{i}}, 
\text{SRE}=\sqrt{\frac{1}{N_{I}}\sum_{i=1}^{N_{I}}\frac{({Y}_{pred}^{i}-{Y}_{gt}^{i})^{2}}{{Y}_{gt}^{i}}}.
\end{gathered}\end{equation}\end{small}

\begin{table}[t]
\small
\begin{center}
\caption{Inference time and parameters. } 
\label{tab:tab3}
\setlength{\tabcolsep}{12pt}
\begin{tabular}{ccc}
  \toprule

  \multirow{1}{*}{Methods} & 
    \multirow{1}{*}{Inference Time(s)} & 
    \multirow{1}{*}{Parameters(M)} 
\\
  \midrule

CLIP-Count* &0.0561$\pm$0.0011& 166.09
\\
CLIP-Count*+ours  &0.0589$\pm$0.0003& 181.29
\\
 \midrule

VLCounter*  &0.0324$\pm$0.0004& 151.35
\\
VLCounter*+ours  &0.0338$\pm$0.0004& 160.49
\\
 \midrule
 
CounTX* &0.0336$\pm$0.0003& 161.08
\\
CounTX*+ours  & 0.0350$\pm$0.0003 & 184.71 
\\
 \midrule
 
CountGD* & 0.1474$\pm$0.0013 & 233.36 
\\
CountGD*+ours & 0.1576$\pm$0.0012 & 238.78 
\\
   \bottomrule
\end{tabular}
\end{center}
\end{table}

\vspace*{-1em}
\subsection{Quantitative Results}
As a plug-and-play framework, we performed integration experiments with all available VLM-based models, i.e., CLIP-Count~\cite{jiang2023clip}, VLCounter~\cite{kang2024vlcounter}, CounTX~\cite{AminiNaieni23}, CountGD~\cite{amini2024countgd}. Given that our framework necessitates reproducing these models prior to integration, we present both quantitative results from their original publications and our reproduced results, denoted as \textbf{Model*}. The integrated experiments are referred to as \textbf{Model*+Ours}.

\begin{table*}[t]
\small
\begin{center}
\caption{
Results on the CARPK and PUCPR+ dataset. The best results are highlighted in bold. } 
\label{tab:tab4}
\setlength{\tabcolsep}{1pt}
\begin{tabular}{ccccccccccc}
  \toprule

  \multirow{2}{*}{Methods} & 
  \multirow{2}{*}{Source} & 
    \multirow{2}{*}{Tuning} & 
  \multicolumn{4}{c}{CARPK}& \multicolumn{4}{c}{PUCPR+}\\
 \cmidrule{4-11}
  & & & MAE$\downarrow$ & RMSE$\downarrow$  & NAE$\downarrow$  &SRE$\downarrow$ & MAE$\downarrow$ & RMSE$\downarrow$  & NAE$\downarrow$ &SRE$\downarrow$ \\
  \midrule

CLIP-Count~\cite{jiang2023clip}  &MM 2023 & VPT &11.96& 16.61 & - & - & -& - & - &- 
\\

CLIP-Count* & - & VPT & 12.03 & 16.04 & 0.42 & 3.11 & 55.82 & 91.23 & 24.17 & 66.78 \\ 

CLIP-Count*+Ours & - & SDVPT & 10.95 {\color{softgreen} $\downarrow$8.98\%} & 14.42 {\color{softgreen} $\downarrow$10.10\%} & 0.34 {\color{softgreen} $\downarrow$19.05\%} & 2.63 {\color{softgreen} $\downarrow$15.43\%} & 48.61 {\color{softgreen} $\downarrow$12.92\%} & 75.09 {\color{softgreen} $\downarrow$17.69\%} & 19.89 {\color{softgreen} $\downarrow$17.71\%} & 56.48 {\color{softgreen} $\downarrow$15.45\%} \\

  \midrule
  
VLCounter~\cite{kang2024vlcounter}  &AAAI 2024& VPT &6.46& 8.68 & -&- & 48.94& 69.08& -&-
\\

VLCounter* & - & VPT & 6.92 & 9.57 & 0.19 & 1.43 & 51.47 & 74.00 & 17.53 & 48.65 \\ 

VLCounter*+Ours & - & SDVPT & 6.31 {\color{softgreen} $\downarrow$8.82\%} & 7.86 {\color{softgreen} $\downarrow$17.87\%} & 0.18 {\color{softgreen} $\downarrow$5.26\%} & 1.41 {\color{softgreen} $\downarrow$1.40\%} & 45.85 {\color{softgreen} $\downarrow$10.92\%} & 62.23 {\color{softgreen} $\downarrow$15.91\%} & 14.72 {\color{softgreen} $\downarrow$16.03\%} & 41.88 {\color{softgreen} $\downarrow$13.92\%} \\

  \midrule

CounTX~\cite{AminiNaieni23}  &BMVC 2023&Full Tuning &11.72&14.86 &-&-& -&-& -&-
\\

CounTX* & - &Full Tuning & 13.13 & 16.92 & 0.29 & 2.22 & 75.24 & 113.44 & 31.17 & 84.72 \\ 

CounTX*+Ours & - & SDVPT & 9.63 {\color{softgreen} $\downarrow$26.66\%} & 13.20 {\color{softgreen} $\downarrow$21.99\%} & 0.28 {\color{softgreen} $\downarrow$3.45\%} & 2.27 {\color{softred} $\uparrow$2.25\%} & 69.54 {\color{softgreen} $\downarrow$7.58\%} & 95.01 {\color{softgreen} $\downarrow$16.25\%} & 25.01 {\color{softgreen} $\downarrow$19.76\%} & 67.77 {\color{softgreen} $\downarrow$20.01\%} \\ 

  \midrule

CountGD~\cite{amini2024countgd}  &NeurIPS 2024& Freeze &3.83& 5.41 & -&-& -& -& -&-
\\
CountGD* & - & Freeze & 4.21 & 5.83 & 0.13 & 0.98 & 24.31 & 32.16 & 0.71 & 5.62 \\ 

CountGD*+Ours & - & SDVPT & \textbf{3.48} {\color{softgreen} $\downarrow$17.34\%} & \textbf{4.53} {\color{softgreen} $\downarrow$22.30\%} & \textbf{0.10} {\color{softgreen} $\downarrow$23.08\%} & \textbf{0.79} {\color{softgreen} $\downarrow$19.39\%} & \textbf{21.20} {\color{softgreen} $\downarrow$12.79\%} & \textbf{24.78} {\color{softgreen} $\downarrow$22.95\%} & \textbf{0.60} {\color{softgreen} $\downarrow$15.49\%} & \textbf{4.37} {\color{softgreen} $\downarrow$22.24\%} \\ 

   \bottomrule
\end{tabular}
\end{center}
\end{table*}
\setlength{\textfloatsep}{5pt}  
\begin{table}[t]
\small
\begin{center}
\caption{ Comparison with other fine-tuning methods on the FSC-147 dataset.} 
\label{tab:tab5}
\setlength{\tabcolsep}{3pt}
\begin{tabular}{ccccccc}
  \toprule

  \multirow{2}{*}{\shortstack{Base \\ Model}} & 
    \multirow{2}{*}{Tuning} & 
    \multirow{2}{*}{\shortstack{Targeted \\ Encoder}} & 
     \multicolumn{2}{c}{Val Set}& \multicolumn{2}{c}{Test Set}\\
 \cmidrule{4-7}
 & & & MAE$\downarrow$ & RMSE$\downarrow$ &MAE$\downarrow$ & RMSE$\downarrow$ \\
  \midrule

CLIP-Count & \multirow{2}{*}{Freeze } & \multirow{2}{*}{- }  & 26.96	& 93.07	&28.84& 120.03
\\

CountGD  &  & & 12.14 & 47.51&12.98& 98.35
\\

 \midrule

CLIP-Count & \multirow{2}{*}{Full Tuning}&  \multirow{2}{*}{Image}& 24.88	& 77.73	&22.51& 131.12
\\

CountGD  & & & 10.83 & 42.81 &14.82 & 112.62
\\

 \midrule

CLIP-Count & \multirow{2}{*}{CoOp} & \multirow{2}{*}{Text} & 27.37	& 90.78	&23.52& 116.33
\\

CountGD  & & &11.74 & 56.30 &15.47 & 118.06
\\

 \midrule

CLIP-Count & \multirow{2}{*}{CoCoOp}& \multirow{2}{*}{Text} &  26.77	& 87.13	&23.91& 120.40
\\

CountGD  &  & &11.33 & 54.98 &15.21 & 115.44
\\

 \midrule

CLIP-Count & \multirow{2}{*}{VPT}& \multirow{2}{*}{Image} & 18.79	& 61.18	&17.78& 106.62
\\

CountGD  & & &10.52 & 52.59&15.09 & 113.31
\\

 \midrule

CLIP-Count & \multirow{2}{*}{SDVPT} & \multirow{2}{*}{Image} & \textbf{16.74}	& \textbf{55.80}	&\textbf{16.92}& \textbf{101.77}
\\

CountGD  & & &\textbf{9.52} &\textbf{ 37.02}& \textbf{11.41}& \textbf{85.50}
\\
   \bottomrule
\end{tabular}
\end{center}
\end{table}

\noindent\textbf{Quantitative Result on FSC-147.} 
As shown in Tab.~\ref{tab:tab1}, our method consistently enhances performance across all integrated models by a significant margin. 
Particularly, when integrating with CountGD~\cite{amini2024countgd}, our method achieves a relative improvement of 15.53\% and 18.27\% in terms of validation MAE and test MAE, respectively, establishing a new state-of-the-art result.
Compared to VPT-based CLIP-Count~\cite{jiang2023clip} and VLCounter~\cite{kang2024vlcounter}, and fully tuned CounTX~\cite{AminiNaieni23}, our method delivers an average NAE improvement of 11.68\%, further validating our effectiveness.

We report the inference time and parameter counts for original models and those integrated with proposed SDVPT in Tab.~\ref{tab:tab3}, validating the low overhead of our framework. 
SDVPT incurs an average time overhead of only 5.10\% and an average parameter increase of just 8.05\%. 
In contrast, compared to the original CounTX~\cite{AminiNaieni23}, the original CountGD~\cite{amini2024countgd} achieves an 18.3\% relative test MAE improvement at the cost of a 338.70\% increase in inference time and a 44.87\% increase in parameters.

\begin{figure}[t]
\centerline{\includegraphics[width=0.45\textwidth]{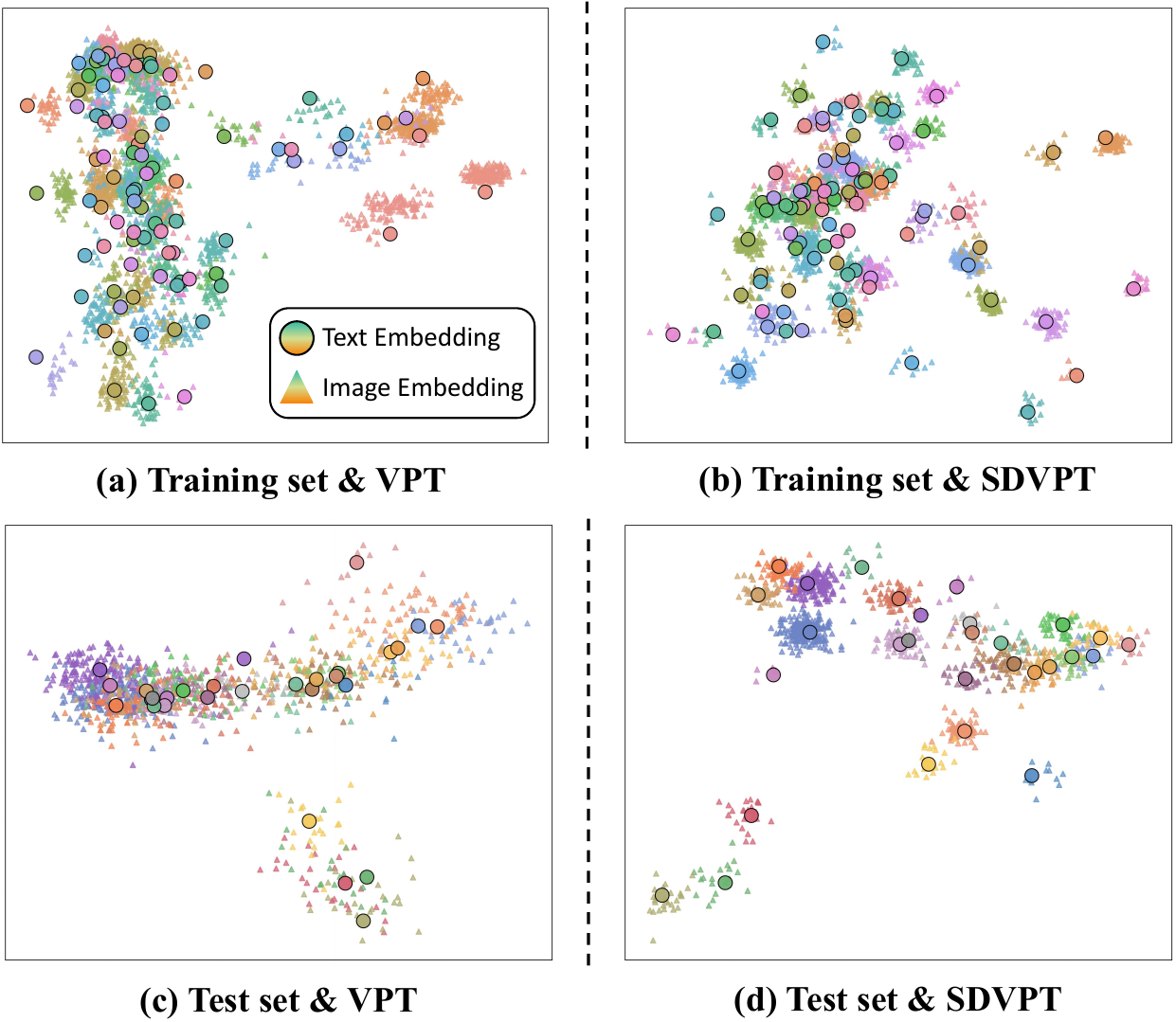}}
\vspace{-1em}
\caption{Joint embedding space of VPT and SDVPT on training and test sets, obtained by dimensionality reduction using Linear Discriminant Analysis (LDA).}
\label{fig:5}
\end{figure}

\begin{figure}[t]
\vspace{-1.5em}
\centerline{\includegraphics[width=0.5\textwidth]{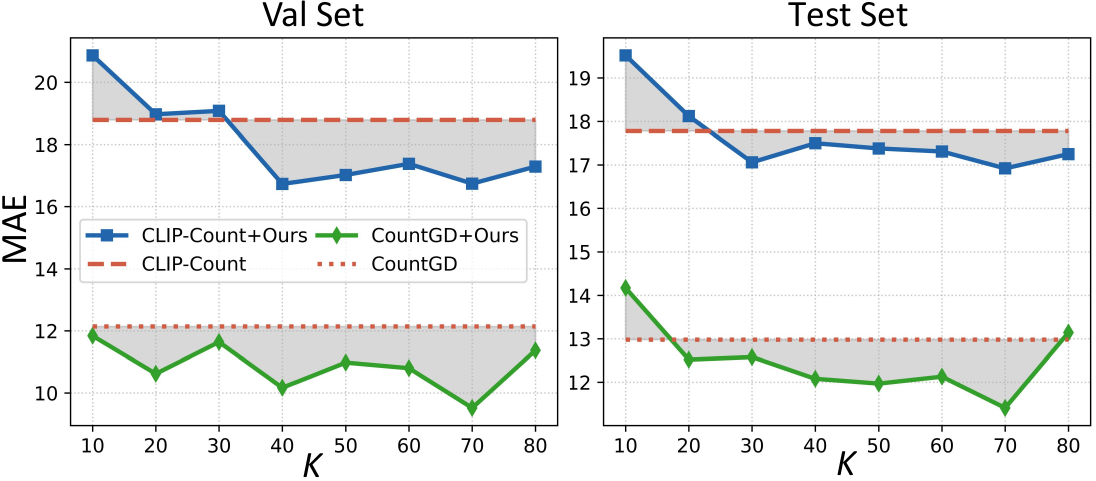}}
\vspace{-1em}
\caption{Ablation study on top-$K$ selection.}
\label{fig:8}
\end{figure}

\begin{figure*}[t]  
\centerline{\includegraphics[width=1\textwidth]{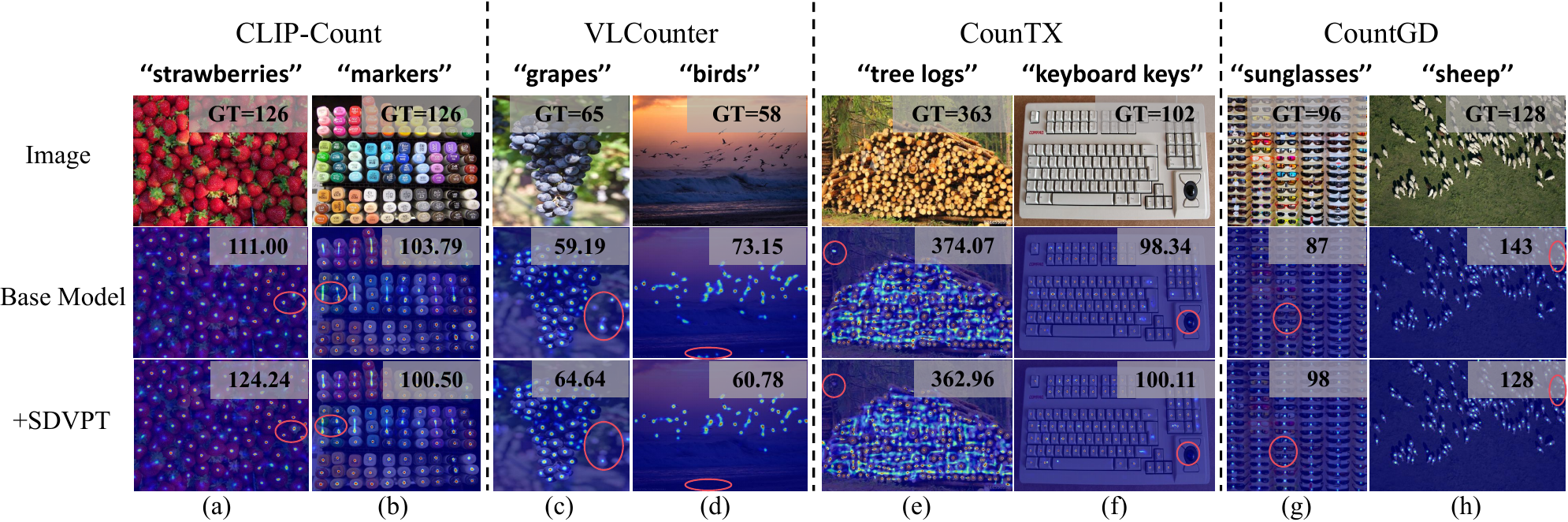}}
\caption{Qualitative comparison of base model and our SDVPT on the FSC-147 dataset. 
}
\label{fig:6}
\end{figure*}

\noindent\textbf{Quantitative Result on CARPK and PUCPR+.} 
Following prior open-world object counting methods~\cite{jiang2023clip,kang2024vlcounter,AminiNaieni23,amini2024countgd}, we performed cross-dataset evaluations on the CARPK and PUCPR+ datasets, as reported in Tab.~\ref{tab:tab4}, to assess the generalization of the proposed SDVPT. 
Specifically, we trained on the FSC-147 dataset and directly evaluated on the test sets of CARPK and PUCPR+. 
SDVPT achieved relative MAE improvements of an average of 14.45\% on the validation set and 11.05\% on the test set across all models, further confirming our method’s effectiveness for unseen categories.

\noindent\textbf{Comparison with Other Fine-Tuning Methods.} 
Using CLIP-Count and CountGD as base models, we evaluated the proposed SDVPT against other prominent fine-tuning methods on the FSC-147 dataset, with results presented in Table~\ref{tab:tab5}. 
CoOp serves as a prompt tuning method for the text encoder, while CoCoOp aligns with SDVPT in addressing generalization to unseen categories, enhancing CoOp with a meta-network for test-time adaptation. 
Our experiments confirm conclusions similar to CounTX~\cite{AminiNaieni23}, showing that despite CoOp and CoCoOp’s excellence in classification tasks as text encoder tuning methods, their effectiveness in open-world counting remains limited, underperforming native full tuning or VPT of the image encoder. 
In contrast, our approach pioneers the use of visual prompts to tackle unseen category generalization, outperforming all selected fine-tuning methods.

With CLIP-Count as the base model, we further visualized the joint embedding space of VPT and the proposed SDVPT across the training and test sets of FSC-147, as illustrated in Fig.~\ref{fig:5}. 
On the training set, SDVPT enforces topological consistency between text and image embeddings for each category, whereas VPT employs a uniform visual prompt across all categories, failing to maintain the per-category text-image consistency.
On the test set, VPT’s disregard for unseen categories results in image embeddings that are more scattered and divergent from their corresponding text embeddings. 
Conversely, our framework effectively extends text-image consistency from the training set to unseen categories, yielding superior prediction accuracy.

\vspace{-0.5em}
\subsection{Ablation Studies}
\label{Ablation_Studies}
\noindent\textbf{Effect of Top-$K$ Selection.} 
We selected CLIP-Count and CountGD as representative models to examine the impact of top-$K$ selection based on the FSC-147 dataset, as depicted in Fig.~\ref{fig:8}. 
On one hand, CLIP-Count and CountGD exhibit similar patterns in top-$K$ selection.
When $K$ ranges from  40 to 80, the proposed SDVPT consistently enhances the performance of base models. However, an overly small $K$ limits diversity, inadequately representing the unseen target class, while an excessively large $K$ introduces undue noise, diminishing the influence of semantically relevant classes.
On the other hand, although the validation and test sets contain entirely different categories, their top-$K$ curves exhibit similar patterns, demonstrating the robustness of our method across diverse categories.

\begin{table}[t]
\setlength{\textfloatsep}{10pt}  
\small
\begin{center}
\caption{Ablation study on each component of SDVPT. } 
\label{tab:tab6}
\setlength{\tabcolsep}{3pt}
\begin{tabular}{cccccccc}
  \toprule

  \multirow{2}{*}{\shortstack{Base \\ Model}} & 
    \multirow{2}{*}{CSPI} & 
    \multirow{2}{*}{TGPR} &
    \multirow{2}{*}{$\mathcal{L}_{\mathrm{recon}}$}&
     \multicolumn{2}{c}{Val Set}& \multicolumn{2}{c}{Test Set}\\
 \cmidrule{5-8}
  & & & & MAE$\downarrow$ & RMSE$\downarrow$ &MAE$\downarrow$ & RMSE$\downarrow$ \\
  \midrule

CLIP-Count & \multirow{2}{*}{\ding{51}} & \multirow{2}{*}{-}  & \multirow{2}{*}{-} & 38.76	& 103.30 &35.03	& 129.40
\\

CountGD  &  &  &  & 14.05	& 50.34	&18.32	& 127.76
\\

  \midrule

  CLIP-Count & \multirow{2}{*}{-} & \multirow{2}{*}{\ding{51}}  & \multirow{2}{*}{-}   & 18.11	&56.83	&17.27	& 105.21
\\

CountGD  &  &  &  & 10.61	&56.66	&12.89	& 109.60
\\

  \midrule

  CLIP-Count & \multirow{2}{*}{\ding{51}} & \multirow{2}{*}{\ding{51}}  & \multirow{2}{*}{-} & 17.59	&57.69	&16.94	& 102.39
\\

CountGD  &  &  &  & 10.31	&44.47	&11.79	& 87.04
\\

  \midrule

  CLIP-Count & \multirow{2}{*}{\ding{51}} & \multirow{2}{*}{\ding{51}}  & \multirow{2}{*}{\ding{51}} &\textbf{16.74}& \textbf{55.80}& \textbf{16.92}& \textbf{101.77}
\\

CountGD  &  &  &  &\textbf{9.52}& \textbf{37.02} & \textbf{11.41}& \textbf{85.50}
\\

   \bottomrule
\end{tabular}
\end{center}
\end{table}

\noindent\textbf{Component Analysis.}
We conducted an ablation study to assess the contribution of each component in SDVPT, as presented in Table~\ref{tab:tab6}. 
Employing CSPI alone yields even inferior results compared to VPT, as it not only disregards text-image consistency for unseen categories but also intensifies overfitting due to its category-specific prompt design. 
However, integrating CSPI with TGPR surpasses TGPR in isolation, underscoring the necessity of pre-training category-specific prompts prior to TGPR application.
Ultimately, the inclusion of $\mathcal{L}_{\mathrm{recon}}$ further boosts counting performance, highlighting the value of explicit structural constraints.
\begin{figure}[t]
\centerline{\includegraphics[width=0.45\textwidth]{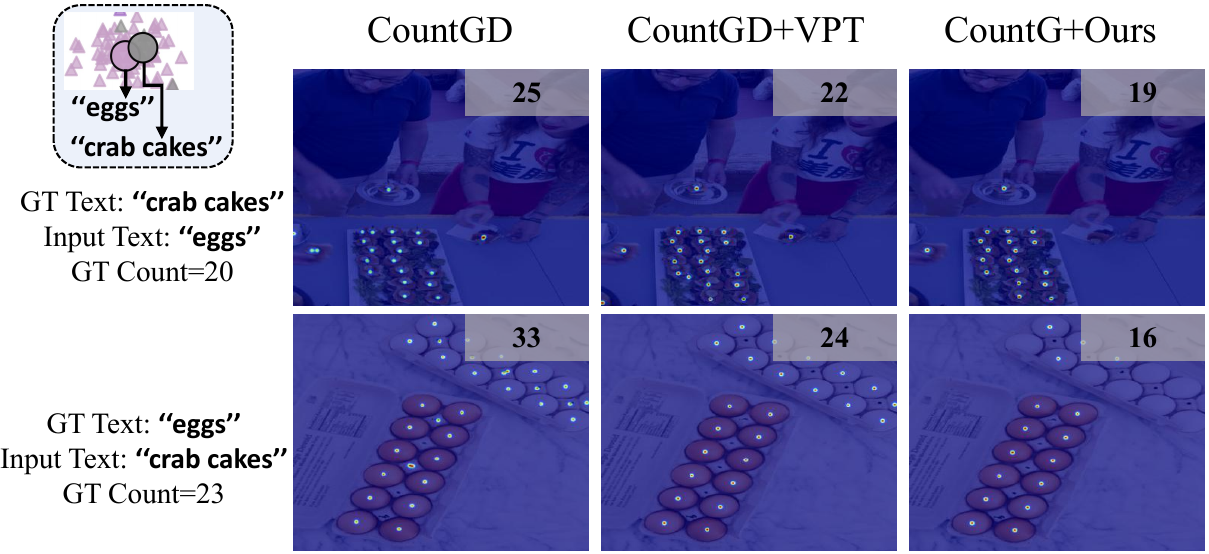}}
\caption{Failures caused by poor separability between two text embeddings.}
\label{fig:7}
\end{figure}

\subsection{Qualitative Results}
We visualize the qualitative results of all base models and those integrated with SDVPT in Fig.~\ref{fig:6}. 
The results in columns (c), (d), (e), (f), and (h) show that our method effectively reduces erroneous predictions in background regions compared to the base models. 
Additionally, the results in columns (a), (b), and (g) indicate that integration with SDVPT enhances the model’s prediction accuracy in regions of the target category.
These observations confirm that the aggregation strategy for visual prompts significantly improves the image-text consistency of the unseen category, thereby reducing predictions in irrelevant regions and yielding more precise predictions in target category regions.

\subsection{Limitations and Future Works}
In Fig.~\ref{fig:5}, we observe that the knowledge in the pre-trained text encoder causes some text embeddings to be closely positioned.
In Fig.~\ref{fig:7}, we selected "eggs" and "crab cakes" whose text embeddings are closely positioned to evaluate the impact of this phenomenon.
For all selected strategies, inputting "eggs" erroneously generates predictions for the "crab cakes" image, while inputting "crab cakes" also yields "eggs" predictions for the image containing only eggs.
In future work, we plan to explore more effective strategies for the text encoder tuning, aiming to reduce inter-category confusion by learning a more separable embedding space.

\section{Conclusion}
In this paper, we demonstrate that considering text-image alignment for unseen categories benefits open-world object counting. 
Specifically, we introduce a plug-and-play visual prompt tuning framework, SDVPT. 
During training, SDVPT utilizes category-specific prompt initialization and topology-guided prompt refinement to transfer training text embedding topologies to a visual prompt set. 
For inference, we dynamically
synthesize the visual prompts for unseen categories to transfer knowledge from the training set, ensuring precise counts. 
Experiments integrating SDVPT with all available open-world counting models across the FSC-147, CARPK, and PUCPR+ datasets confirm its effectiveness.

\bibliographystyle{ACM-Reference-Format}
\bibliography{sample-base}

\appendix
\clearpage
\section{Additional Implementation Details}

When integrated with base models, we adopt the text description formats and training category counts from their original implementation details as follows:

\begin{itemize}
\item CLIP-Count employs the original category names of the FSC-147 dataset as text descriptions, denoted as "\{ \}s", where "\{ \}" represents the singular form of the category name (e.g., "apple"), encompassing 89 training categories.
\item VLCounter incorporates the original category names into 11 templates, also utilizing 89 training categories:
\begin{itemize}
\item "A photo of a number of \{ \}s."
\item "A photo of a number of small \{ \}s."
\item "A photo of a number of medium \{ \}s."
\item "A photo of a number of large \{ \}s."
\item "There is a photo of a number of \{ \}s."
\item "There is a photo of a number of small \{ \}s."
\item "There is a photo of a number of medium \{ \}s."
\item "There is a photo of a number of large \{ \}s."
\item "A number of \{ \}s in the scene."
\item "A photo of a number of \{ \}s in the scene."
\item "There are a number of \{ \}s in the scene."
\end{itemize}
\item CounTX introduces a list, FSC-147-D, to redefine the FSC-147 training category names. The text descriptions are formatted as "the \{ \}s", and the training categories are expanded to 123.
\item CountGD refines FSC-147-D, reducing the training categories to 90, while adopting text descriptions in the form of "\{ \}".
\end{itemize}
As analyzed in Section~\ref{Ablation_Studies} regarding the top-$K$ selection, due to CounTX employing a larger number of training categories, we utilize a greater $K$ (80) for CounTX compared to other base models, as shown in Tab.~\ref{tab:2}.

\section{Further Comparison with VPT}
\begin{figure}[h]
\centerline{\includegraphics[width=0.5\textwidth]{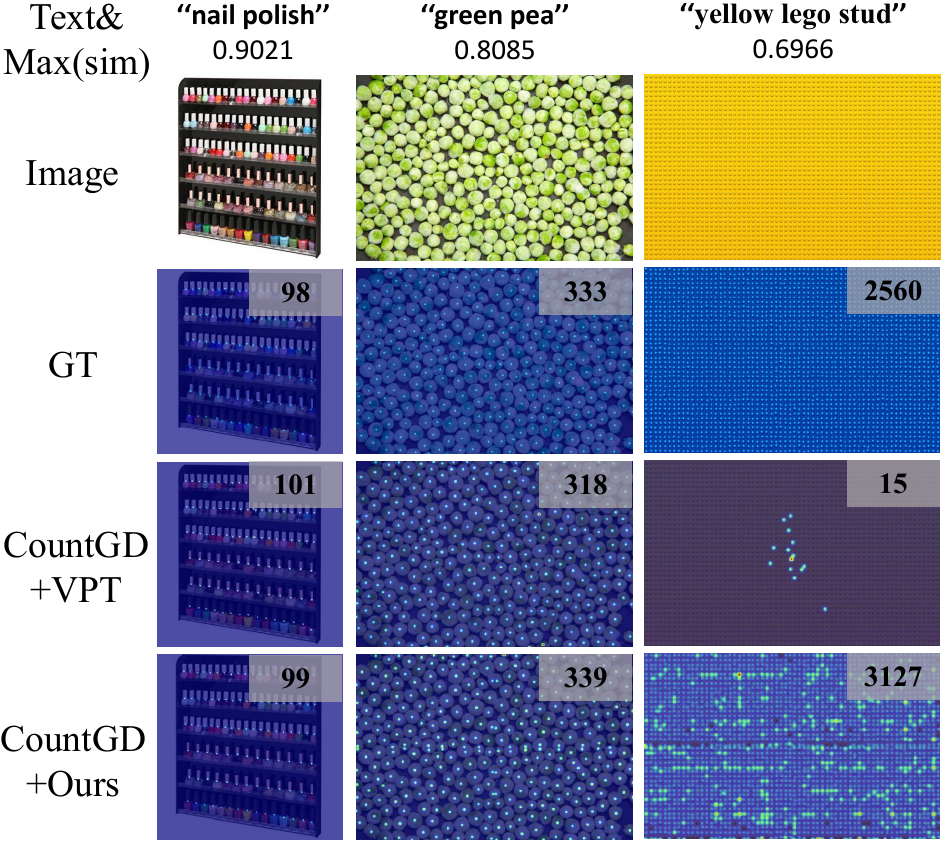}}
\caption{Qualitative comparison of VPT and our SDVPT on the FSC-147 dataset. MAX(sim) denotes the maximum cosine similarity between the input text and the text of all training set categories. The counting result or GT is displayed on the top right of the corresponding image.}
\label{fig_appendix:1}
\end{figure}

As shown in Fig.~\ref{fig_appendix:1}, we present qualitative results to evaluate the generalization capabilities of VPT and our SDVPT across unseen categories with varying degrees of unfamiliarity.
The first column shows that VPT can yield relatively accurate predictions when unseen categories closely resemble training categories.
However, as this similarity diminishes, the errors increase, particularly in the third column where the MAX(sim) is only 0.6.
We hypothesize that this is due to substantial differences with training categories, which prevent VPT from ensuring text-image alignment for that category.
In contrast, our method not only achieves greater prediction accuracy for "nail polish" and "green pea", which are relatively similar to training categories, but also maintains robust performance for the markedly dissimilar "yellow lego stud" category.

\section{Further Ablations}
In addition to the $L_2$ distance, we also explored cosine similarity as a distance metric between the fused prompt and the category-specific prompt for the $\mathcal{L}_{\text{recon}}$. 
Ablation studies conducted on FSC-147 with CLIP-Count as the base model are presented in Tab.~\ref{tab:tab_appendix_1}.
Employing the L2 distance as the metric yields superior performance. 
We hypothesize that this may be because cosine similarity focuses solely on directional information, while the L2 distance considers both direction and vector length, preserving richer geometric structure information.
 
\begin{table}[h]
\setlength{\textfloatsep}{12pt}  
\begin{center}
\caption{Ablation study on the distance metric in $\mathcal{L}_{\text{recon}}$ based on the CLIP-Count.} 
\label{tab:tab_appendix_1}
\setlength{\tabcolsep}{3pt}
\begin{tabular}{cccccccc}
  \toprule
  \multirow{1}{*}{Metric} & 
     \multirow{1}{*}{Val MAE}& 
     \multirow{1}{*}{Val RMSE}& 
     \multirow{1}{*}{Test MAE}& 
     \multirow{1}{*}{Test RMSE}
     \\
  \midrule
Cosine  & 17.44	& 59.46 &18.03	& 102.03
\\
$L_2$ & \textbf{16.74}	& \textbf{55.80}	&\textbf{16.92}& \textbf{101.77}
\\
   \bottomrule
\end{tabular}
\end{center}
\end{table}

\section{Additional Qualitative Results}
\begin{figure*}[t]
\centerline{\includegraphics[width=1\textwidth]{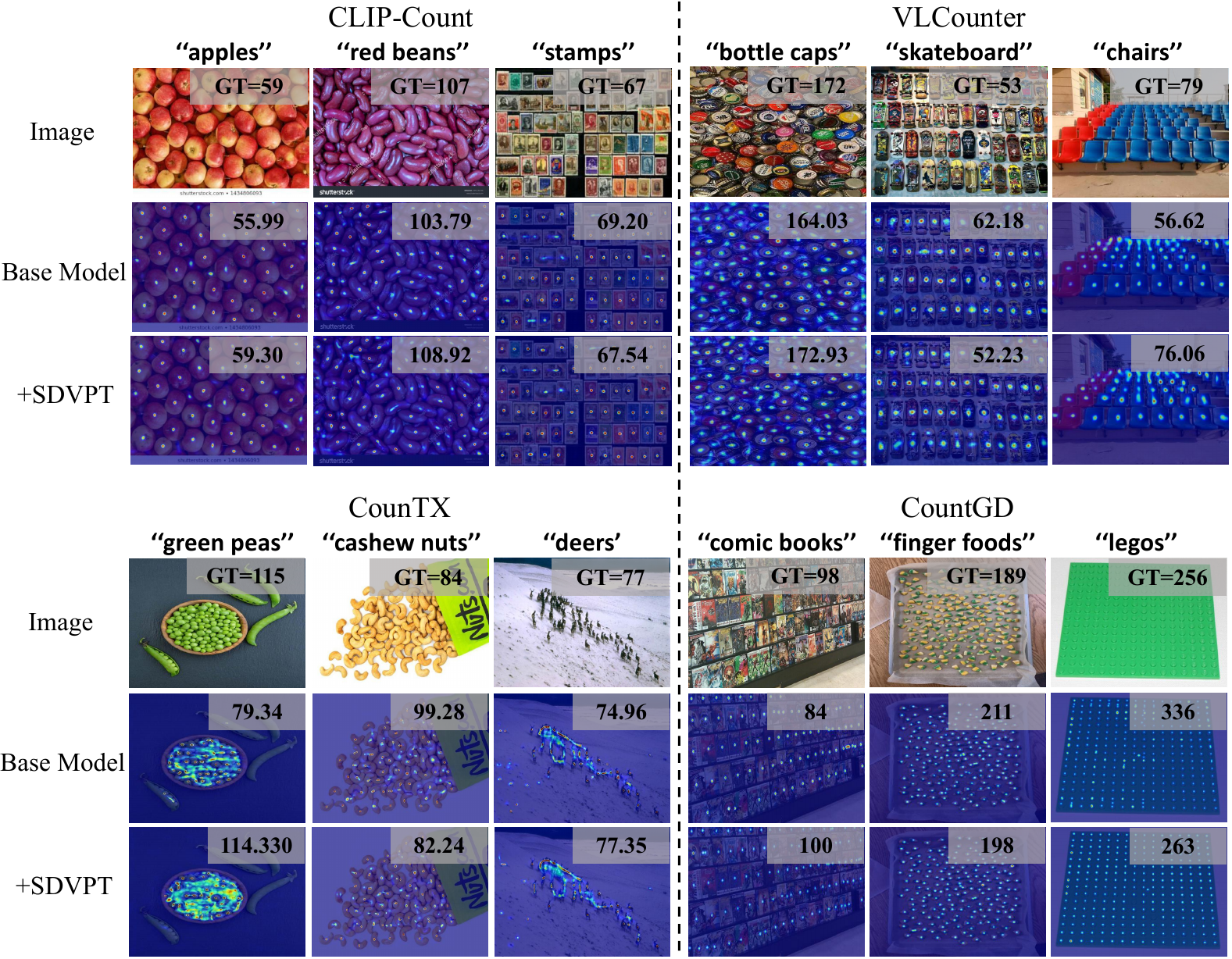}}
\caption{Additional qualitative results. The counting result or GT is displayed on the top right of the corresponding image.}
\label{fig_appendix:2}
\end{figure*}

Beyond the qualitative results in the manuscript, we provide additional results in Fig.~\ref{fig_appendix:2}. Our method enhances the counting performance of various base models across diverse unseen categories, demonstrating the generality of the proposed SDVPT.

\end{document}